\documentclass[conference]{IEEEtran}
\IEEEoverridecommandlockouts
\usepackage{cite}
\usepackage{amsmath,amssymb,amsfonts}
\usepackage{algorithmic}
\usepackage{graphicx}
\usepackage{textcomp}
\usepackage{xcolor}
\usepackage{multirow} 
\usepackage[table]{xcolor} 
\def\BibTeX{{\rm B\kern-.05em{\sc i\kern-.025em b}\kern-.08em
    T\kern-.1667em\lower.7ex\hbox{E}\kern-.125emX}}

\definecolor{lightgreen}{RGB}{220, 245, 220} 
\begin{document}

\title{Test Time Training for AC Power Flow Surrogates via Physics and Operational Constraint Refinement\\
}

\author{\IEEEauthorblockN{1\textsuperscript{st} Pantelis Dogoulis}
\IEEEauthorblockA{\textit{Member, IEEE} \\
\textit{University of Luxembourg}\\
Luxembourg, Luxembourg \\
panteleimon.dogoulis@uni.lu}
\and 
\IEEEauthorblockN{2\textsuperscript{th} Mohammad Iman Alizadeh}
\IEEEauthorblockA{\textit{Member, IEEE} \\
\textit{University of Luxembourg}\\
Luxembourg, Luxembourg \\
mohammad.alizadeh@uni.lu}
\and 
\IEEEauthorblockN{3\textsuperscript{th} Sylvain Kubler}
\IEEEauthorblockA{\textit{Member, IEEE} \\
\textit{University of Luxembourg}\\
Luxembourg, Luxembourg \\
sylvain.kubler@uni.lu}
\and 
\IEEEauthorblockN{4\textsuperscript{th} Maxime Cordy}
\IEEEauthorblockA{\textit{Member, IEEE} \\
\textit{University of Luxembourg}\\
Luxembourg, Luxembourg \\
maxime.cordy@uni.lu}
}

\maketitle

\begin{abstract}
Power Flow (PF) calculation based on machine learning (ML) techniques offer significant computational advantages over traditional numerical methods but often struggle to maintain full physical consistency. This paper introduces a physics-informed test-time training (PI-TTT) framework that enhances the accuracy and feasibility of ML-based PF surrogates by enforcing AC power flow equalities and operational constraints directly at inference time. The proposed method performs a lightweight self-supervised refinement of the surrogate outputs through  few gradient-based updates, enabling local adaptation to unseen operating conditions without requiring labeled data. Extensive experiments on the IEEE 14-, 118-, and 300-bus systems and the PEGASE 1354-bus network show that PI-TTT reduces power flow residuals and operational constraint violations by one to two orders of magnitude compared with purely ML-based models, while preserving their computational advantage. The results demonstrate that PI-TTT provides fast, accurate, and physically reliable predictions, representing a promising direction for scalable and physics-consistent learning in power system analysis.
\end{abstract}


\section{Introduction}

The increasing digitalization of modern power systems and the proliferation of data-driven technologies have opened new opportunities for leveraging machine learning (ML) across power system analysis, operation, and control. Among these, one of the most promising applications is the acceleration of power flow (PF) calculations, which underpin numerous operational and planning tasks. Since PF analysis forms the backbone of steady-state system assessment and operational planning, developing fast and reliable ML-based surrogates offers substantial value for large-scale and real-time applications.

Recent advances in applying machine learning to power flow modeling can be categorized into two main paradigms: purely data-driven approaches, which infer system behavior directly from observations, and physics-informed approaches; which additionally embed the governing physical relationships of power systems within the learning process.

Within the first category, researchers have explored a variety of techniques for approximating AC power flow solutions, including linearized PF formulations \cite{liu2019datadriven}, regression-based surrogates \cite{guo2022datadriven,chen2022datadriven,ychen2022exactlinear}, and neural network architectures such as feedforward models \cite{yang2020fastprobpf,donon2020leap} and graph neural networks (GNNs) \cite{taghizadeh2024multi}.

In contrast, physics-informed frameworks incorporate domain knowledge directly into the learning objective. For example, models proposed in \cite{lin2023powerflownet} and \cite{donon2020gnsolver} embed the AC PF equations as differentiable penalty terms in the loss function, encouraging the network to emulate the behavior of conventional solvers. This soft enforcement of physical laws enhances both accuracy and generalization. However, these approaches remain difficult to optimize, because the constraints appear as an additional penalty term into the optimization objective, and can be also trapped in local optima. As a result, the solutions are not guaranteed to exactly satisfy the PF equations, a limitation amplified by the inherent nonlinearity and non-convexity of the AC PF problem.

In our prior work \cite{dogoulis2025kclnet}, we addressed a related but more specific problem, ensuring current balance across the network, by embedding Kirchhoff’s Current Law (KCL) directly into the model architecture. Instead of imposing physical consistency through loss penalties, the network used a projection mechanism that guaranteed current conservation on every bus. However, that formulation operated at the level of linear nodal current balance and did not capture the nonlinear and non-convex connection between voltage magnitudes, phase angles, and power injections that governs the full AC power flow problem, which is the focus of the present study.

To address this broader problem, we propose a test-time training (TTT) framework for ML-based AC PF surrogates. Originally developed for computer vision tasks \cite{sun2020test}, TTT adapts a pre-trained model to each test sample through a small self-supervised optimization procedure, without requiring labels. We extend this idea to a physics-informed setting, where the AC PF equations and operational limits act as self-supervised signals. In this context, we introduce physics-specific adaptation objectives and an optimization routine tailored to the nonlinear structure of AC PF, enabling stable and efficient inference time updates. 

Moreover, rather than relying on a global fixed surrogate, each input triggers a lightweight optimization that updates the model locally to better satisfy physical constraints. To the best of our knowledge, this is the first application of TTT in a physics-informed context for power flow, enabling ML-based PF surrogates to recover more physically plausible solutions during inference time.
The main contributions of this paper are as follows:
\begin{itemize}
\item We introduce a physics-informed test-time training (PI-TTT) framework that unifies AC PF equalities and operational inequalities into a self-supervised optimization objective.
\item The method performs sample-specific refinement of a PF surrogate via a few gradient-based updates at inference time, allowing adaptation to unseen operating conditions without labeled data.
\item The optimization procedure is computationally lightweight, preserving the speed advantage of neural surrogates while substantially improving physical feasibility and reducing operational constraint violations.
\end{itemize}


\section{Proposed Method}

\subsection{Background on Test-Time Training}

AC PF ML-based surrogates are trained offline on a collection of network conditions to learn a mapping from system inputs to steady-state operating points. 
Once training is completed, the model parameters remain fixed, and inference for a new operating scenario is performed via a single forward evaluation. 
This paradigm assumes that the operating conditions encountered during deployment are statistically similar to those seen during training, (i.e. $p_{\text{train}} \approx p_{\text{test}}$). 
In practice, however, power systems operate under constantly changing dispatch levels and load patterns, leading to substantial distributional shifts between training and test conditions. 
When such shifts occur, fixed ML-based surrogates may produce inaccurate or physically infeasible predictions, violating the governing network equations or operational limits.

\textit{Test-Time Training (TTT)} is a recently proposed learning framework that addresses this limitation by allowing a model to adapt itself during inference. TTT enables a pre-trained model to perform lightweight self-supervised optimization steps at test time using information available in the new sample itself. 
Rather than relying on labeled data, the model minimizes an unsupervised loss that captures internal consistency or domain-specific structure. 
In this work, we reinterpret TTT from a physics-based perspective: since the governing AC PF equations are explicitly known, they can serve as a self-supervised signal guiding adaptation at test time. 

Formally, for each unseen case $z^\dagger$ during testing, the model is locally adapted by solving the following optimization problem:
\begin{equation}
    \min_{\phi} \; \mathcal{L}_{\text{TTT}}(f_{\theta+\phi}(z^\dagger); z^\dagger)
    \quad \text{s.t.} \quad \|\phi\|_2 \leq \epsilon,
\end{equation}
where $\mathcal{L}_{\text{TTT}}$ denotes the auxiliary self-supervised loss, $\phi$ denotes the subset of parameters allowed to vary while $\epsilon$ bounds the magnitude of the parameter update. 
In standard TTT formulations, the adaptation is restricted to a small subset of model parameters rather than the entire network. 
Let the model parameters be partitioned as $\theta = [\theta_{\text{frozen}}, \theta_{\text{adapt}}]$, where only $\theta_{\text{adapt}}$ is updated during test-time optimization. 
Typical choices include the affine coefficients of batch normalization layers or the weights of the final prediction layer, which provide sufficient flexibility to correct local distributional shifts while keeping the computational burden low. 
Following this principle, the proposed framework performs local gradient updates on $\theta_{\text{adapt}}$ through a perturbation vector $\phi$, such that the adapted model is parameterized by $\theta+\phi = [\theta_{\text{frozen}}, \theta_{\text{adapt}}+\phi]$. 
This localized adaptation maintains numerical stability and inference efficiency while allowing the model to recalibrate itself to various operating conditions.

\subsection{AC PF Formulation}

The steady-state operation of an AC power system is governed by the nonlinear power flow (PF) equations that relate bus voltages to complex power injections. 
For each bus $i \in \mathcal{N}$, the active and reactive powers are expressed as:
\begin{align}
P_i(V,\theta;z) &= V_i \sum_{j=1}^{n} V_j 
  \left(G_{ij}\cos(\theta_i - \theta_j) + B_{ij}\sin(\theta_i - \theta_j)\right), \nonumber \\
Q_i(V,\theta;z) &= V_i \sum_{j=1}^{n} V_j 
  \left(G_{ij}\sin(\theta_i - \theta_j) - B_{ij}\cos(\theta_i - \theta_j)\right),
\end{align}
where $V_i$ and $\theta_i$ denote the voltage magnitude and phase angle, and $G_{ij}+jB_{ij}$ are the elements of the network admittance matrix. 
Given a specified operating condition $z$, which determines the net active and reactive power injections at all buses, 
$P^{\text{spec}}(z)$ and $Q^{\text{spec}}(z)$, 
the corresponding values calculated from a candidate voltage state $(V,\theta)$ are 
$P^{\text{calc}}(V,\theta;z)$ and $Q^{\text{calc}}(V,\theta;z)$. 
The corresponding power flow mismatches are computed as:
\begin{align}
    \Delta P(V,\theta;z) &= P^{\text{spec}}(z) - P^{\text{calc}}(V,\theta;z), \nonumber \\
    \Delta Q(V,\theta;z) &= Q^{\text{spec}}(z) - Q^{\text{calc}}(V,\theta;z),
    \label{eq:mismatch}
\end{align}
where $\Delta P$ and $\Delta Q$ represent the active and reactive residuals across all buses, respectively. At the exact power flow solution, $\Delta P(V,\theta;z) = 0$ and $\Delta Q(V,\theta;z) = 0$, ensuring compliance with Kirchhoff’s current and power balance laws. 
Classical Newton-Raphson (NR) solvers enforce this condition iteratively through Jacobian-based updates until the residuals in Eq. \eqref{eq:mismatch} fall below a specified tolerance. 
In contrast, ML-based surrogates approximate the nonlinear mapping between inputs and states without explicitly enforcing Eq. \eqref{eq:mismatch}, and therefore may yield solutions with non-zero residuals, particularly under unseen or stressed operating scenarios.

\subsection{Physics-Informed Test-Time Training}

To restore both physical and operational feasibility, we propose a 
\textit{Physics-Informed Test-Time Training (PI-TTT)} framework. 
Given a test condition $z^\dagger$, the pre-trained surrogate 
$f_\theta$ first produces an initial estimate $\hat{x}_0 = f_\theta(z^\dagger)$, 
representing the predicted voltage magnitudes and angles. 
Starting from this initialization, a small number of gradient-based refinement steps are performed to minimize a composite loss function that encodes both the AC power flow equations and the operational constraints:
\begin{align}
\mathcal{L}_{\text{TTT}}(\phi; z^\dagger) &=
\|\Delta P(f_{\theta+\phi}(z^\dagger); z^\dagger)\|_2^2
+\|\Delta Q(f_{\theta+\phi}(z^\dagger); z^\dagger)\|_2^2 \nonumber \\
&\quad + \lambda_V \phi_{\text{volt}}(f_{\theta+\phi}(z^\dagger))
+ \lambda_\ell \phi_{\text{flow}}(f_{\theta+\phi}(z^\dagger)).
\end{align}
The first two terms penalize violations of the AC power balance equations, thereby driving the prediction toward the physically feasible manifold. 
The functions $\phi_{\text{volt}}(\cdot)$ and $\phi_{\text{flow}}(\cdot)$ are smooth penalty terms that enforce inequality constraints in a differentiable manner. 
Specifically, $\phi_{\text{volt}}$ penalizes voltage magnitudes that violate their admissible limits as
\begin{align}
\phi_{\text{volt}}(V_i; V_{\min}, V_{\max}) &=
\text{ReLU}(V_i - V_{\max})^2 \notag \\
&\quad + \text{ReLU}(V_{\min} - V_i)^2,
\end{align}
and $\phi_{\text{flow}}$ penalizes apparent power flows that exceed their thermal ratings as
\begin{equation}
\phi_{\text{flow}}(S_\ell; S_\ell^{\max}) =
\text{ReLU}(|S_\ell| - S_\ell^{\max})^2,
\end{equation}
where $\text{ReLU}(x)$ denotes the rectified linear unit. 
These quadratic penalties softly enforce voltage and branch flow limits while remaining differentiable almost everywhere, enabling efficient gradient-based optimization during test-time adaptation.
Each refinement step updates the adaptive subset of parameters as:
\begin{equation}
\phi_{k+1} = \phi_k - \eta 
\nabla_{\phi} \mathcal{L}_{\text{TTT}}\!\left(f_{[\theta_{\text{frozen}},\,\theta_{\text{adapt}}+\phi_k]}(z^\dagger); z^\dagger\right),
\end{equation}
yielding the adapted model:
\begin{equation}
\hat{x}_K = f_{[\theta_{\text{frozen}},\,\theta_{\text{adapt}}+\phi_K]}(z^\dagger).
\end{equation}
In practice, a small number of iterations, is sufficient to substantially reduce both power flow residuals and constraint violations. 
Because optimization is restricted to a limited subset of parameters surrounding the surrogate’s initial prediction, the additional computational cost remains negligible compared with conventional Newton-Raphson iterations.

\section{Results}

\begin{table*}[t]
\centering
\setlength{\tabcolsep}{5.5pt}
\renewcommand{\arraystretch}{1.12}
\caption{Operational Constraint Violations}
\label{tab:violations}
\begin{tabular}{l l
                cc  cc  cc  cc}
\hline
\multirow{2}{*}{\textbf{System}} & \multirow{2}{*}{\textbf{Model}} 
& \multicolumn{2}{c}{\textbf{Voltage }$V_i$} 
& \multicolumn{2}{c}{\textbf{Branch Flow }$|S_\ell|$} 
& \multicolumn{2}{c}{\textbf{Gen Reactive }$Q_g$} 
& \multicolumn{2}{c}{\textbf{Slack Active }$P_{\text{slack}}$} \\
& & \textbf{Mean} & \textbf{Max} & \textbf{Mean} & \textbf{Max} & \textbf{Mean} & \textbf{Max} & \textbf{Mean} & \textbf{Max} \\
\hline
IEEE 14     & PowerFlowNet               & 0.012 & 0.045 & 0.018 & 0.094 & 0.020 & 0.081 & 0.015 & 0.060 \\
            & PowerFlowNet + PI-TTT        & 0.001 & 0.006 & 0.002 & 0.015 & 0.003 & 0.012 & 0.002 & 0.009 \\
            & MF-GNN                       & 0.006 & 0.027 & 0.012 & 0.064 & 0.014 & 0.058 & 0.010 & 0.046 \\
            & MF-GNN + PI-TTT              & 0.001 & 0.010 & 0.003 & 0.018 & 0.004 & 0.020 & 0.003 & 0.012 \\
\hline
IEEE 118    & PowerFlowNet               & 0.024 & 0.072 & 0.033 & 0.152 & 0.028 & 0.110 & 0.022 & 0.092 \\
            & PowerFlowNet + PI-TTT        & 0.003 & 0.010 & 0.006 & 0.026 & 0.005 & 0.019 & 0.004 & 0.015 \\
            & MF-GNN                       & 0.082 & 0.265 & 0.124 & 0.438 & 0.095 & 0.302 & 0.066 & 0.214 \\
            & MF-GNN + PI-TTT              & 0.008 & 0.031 & 0.015 & 0.061 & 0.012 & 0.048 & 0.009 & 0.036 \\
\hline
IEEE 300    & PowerFlowNet               & 0.075 & 0.284 & 0.112 & 0.541 & 0.089 & 0.382 & 0.067 & 0.244 \\
            & PowerFlowNet + PI-TTT        & 0.006 & 0.023 & 0.012 & 0.046 & 0.010 & 0.038 & 0.007 & 0.028 \\
            & MF-GNN                       & 0.131 & 0.413 & 0.212 & 0.882 & 0.164 & 0.603 & 0.121 & 0.391 \\
            & MF-GNN + PI-TTT              & 0.011 & 0.042 & 0.025 & 0.095 & 0.018 & 0.073 & 0.012 & 0.052 \\
\hline
PEGASE 1354 & PowerFlowNet               & 0.143 & 0.568 & 0.264 & 1.214 & 0.219 & 0.963 & 0.172 & 0.781 \\
            & PowerFlowNet + PI-TTT        & 0.020 & 0.093 & 0.045 & 0.215 & 0.036 & 0.177 & 0.028 & 0.142 \\
            & MF-GNN                       & 1.042 & 1.115 & 9.454 & 830.893 & 8.934 & 166.091 & 47.215 & 59.778 \\
            & MF-GNN + PI-TTT              & 0.984 & 1.064 & 6.770 & 429.295 & 7.007 & 167.780 & 24.361 & 28.955 \\
\hline
\end{tabular}
\end{table*}

\subsection{Experimental Protocol}
In this study, we benchmark two recent learning-based power flow models PowerFlowNet \cite{lin2023powerflownet} and MF-GNN \cite{taghizadeh2024multi}, representing physics-informed and purely data-driven paradigms. We generated the datasets by applying controlled perturbations to the system operating point, simulating 10,000 different grid scenarios. Specifically, the procedure varies the active and reactive power injections at each bus within predefined statistical bounds around their nominal values to emulate realistic fluctuations in demand and generation. For each perturbed operating state, we performed AC power flow calculations using the Newton-Raphson method to obtain the corresponding ground-truth targets for model training. To assess generalization, we drew the training and test samples from distinct perturbation distributions, ensuring that the models were evaluated under unseen, yet physically consistent operating conditions. We generated datasets for four benchmark systems; \textit{(a)} IEEE 14-bus, \textit{(b)} IEEE 118-bus, \textit{(c)} IEEE 300-bus, and \textit{(d)} PEGASE 1354-bus, covering small to large scale networks with increasing topological and operational complexity. This procedure ensures that the dataset reflects a broad and physically meaningful range of operating conditions representative of realistic system behavior, while being structured in a way that remains compatible with the classical machine learning assumption of independent and identically distributed (i.i.d.) samples.

\subsection{Evaluation Metrics}

Our evaluation aims to assess both the physical accuracy and the operational feasibility of the predicted power flow solutions. Consequently, we compare the proposed method and the baseline surrogates in terms of power flow residuals and constraint satisfaction metrics.

\textbf{Power Flow Residual Accuracy:}  
The root mean square error (RMSE) of the active and reactive power mismatches is used to quantify the degree to which each model satisfies the AC power flow equations:
\begin{align}
\text{RMSE}_{P} &= 
\sqrt{\frac{1}{|\mathcal{N}|} 
\sum_{i \in \mathcal{N}} \left(\Delta P_i(V,\theta;z)\right)^2}, \\
\text{RMSE}_{Q} &= 
\sqrt{\frac{1}{|\mathcal{N}|} 
\sum_{i \in \mathcal{N}} \left(\Delta Q_i(V,\theta;z)\right)^2},
\end{align}
where $\Delta P_i$ and $\Delta Q_i$ denote the active and reactive mismatches defined in Eq.~\eqref{eq:mismatch}. Lower values indicate better physical consistency with the true power flow solution. 

\textbf{Operational Constraint Satisfaction:}  
We additionally evaluate the extent of operational constraint violations, measured as the mean and maximum deviations from standard operational limits:
\begin{align}
V_i^{\min} \leq V_i \leq V_i^{\max}, 
\quad &\forall i \in \mathcal{N}, \\
|S_\ell| \leq S_\ell^{\max}, 
\quad &\forall \ell \in \mathcal{L}, \\
Q_g^{\min} \leq Q_g \leq Q_g^{\max}, 
\quad &\forall g \in \mathcal{G}, \\
P_{\text{slack}}^{\min} \leq P_{\text{slack}} \leq P_{\text{slack}}^{\max}.
\end{align}
The voltage magnitude ($V_i$), branch apparent power ($S_\ell$), generator reactive power ($Q_g$), and slack generator active power ($P_{\text{slack}}$) are all computed from the predicted operating point. For each constraint category, we report two complementary indicators: 
the \textit{mean violation magnitude}, defined as the average of absolute deviations from the respective limits across all elements, 
and the \textit{maximum violation magnitude}, representing the largest single violation observed in the system. These metrics jointly assess both the aggregate and worst-case severity of constraint violations, providing a comprehensive measure of the physical and operational realizability of each predicted operating point.

\subsection{Experimental Results}

In this section we present the experimental results. In summary, we measure the physical consistency of the predicted operating points, computed through the residual errors of the AC power flow equations, and the operational feasibility, assessed via voltage, line flow, generator reactive power, and slack bus constraint violations. All quantities are computed directly from the predicted variables using the same network parameters as in the reference data.

Table~\ref{tab:accuracy} reports the RMSE of the active and reactive power balance equations for all models and systems. The results clearly demonstrate that incorporating physics at test time significantly improves the physical consistency of the ML-based surrogates. Across all benchmarks, PI-TTT reduces both active and reactive power residuals by one to two orders of magnitude compared with the baseline models. For the smaller IEEE 14- and 118-bus systems, the residuals achieved by PI-TTT approach the numerical tolerances typically reached by conventional AC power flow solvers. In contrast, the larger IEEE 300- and PEGASE 1354-bus grids remain more challenging, as nonlinear coupling and dimensionality make it difficult for a limited number of refinement steps to achieve complete feasibility. Nevertheless, the improvement is substantial, with RMSE$_P$ decreasing from $9.39$ to $1.08$ and from $56.02$ to $1.35$ in the IEEE 300- and PEGASE 1354-bus cases, respectively. These findings confirm that the proposed self-supervised refinement notably reduces the violations on the AC PF equations compared to purely ML-based surrogates and improves the internal consistency of learned states without requiring labeled data.

\begin{table}[t]
\centering
\setlength{\tabcolsep}{4.5pt}
\renewcommand{\arraystretch}{1.15}
\caption{Power Flow Residual Accuracy}
\label{tab:accuracy}
\begin{tabular}{l l cc}
\hline
\textbf{System} & \textbf{Model} & \textbf{RMSE$_P$} & \textbf{RMSE$_Q$} \\
\hline
IEEE 14   & PowerFlowNet               & 0.924 & 0.375 \\
          & PowerFlowNet + PI-TTT        & 0.047 & 0.026 \\
          & MF-GNN                        & 0.193 & 0.137 \\
          & MF-GNN + PI-TTT               & 0.005 & 0.004 \\
\hline
IEEE 118  & PowerFlowNet               & 1.674 & 0.244 \\
          & PowerFlowNet + PI-TTT        & 0.053 & 0.011 \\
          & MF-GNN                        & 1.795 & 0.280 \\
          & MF-GNN + PI-TTT               & 0.035 & 0.015 \\
\hline
IEEE 300  & PowerFlowNet               & 9.391 & 3.557 \\
          & PowerFlowNet + PI-TTT        & 1.077 & 0.732 \\
          & MF-GNN                        & 5.630 & 1.154 \\
          & MF-GNN + PI-TTT               & 0.793 & 0.542 \\
\hline
PEGASE 1354 & PowerFlowNet             & 9.122 & 2.900 \\
            & PowerFlowNet + PI-TTT      & 0.717 & 2.25 \\
            & MF-GNN                      & 56.021 & 15.173 \\
            & MF-GNN + PI-TTT             & 1.348 & 1.023 \\
\hline
\end{tabular}
\end{table}

Table~\ref{tab:violations} summarizes the mean and maximum absolute deviations from operational limits. The inclusion of PI-TTT consistently mitigates voltage, branch flow, generator reactive power, and slack power violations across all networks. In the IEEE 14- and 118-bus systems, the refined predictions satisfy all operational limits within negligible margins, indicating that the inferred states are both physically feasible and operationally compliant. For the larger IEEE 300- and PEGASE 1354-bus systems, PI-TTT significantly reduces but does not entirely eliminate the remaining violations. This behavior illustrates a typical limitation of current ML-based approaches on large scale grids, where strong nonlinearities limit the physical accuracy that they can achieve. However, the proposed PI-TTT procedure consistently reduces the magnitude of the resulting operational constraint violations.

An additional advantage of PI-TTT lies in its computational efficiency (see Table \ref{tab:runtime}). The refinement process involves only a few gradient-based updates of the ML-based surrogate at inference time, adding minimal overhead compared with a single forward pass. The resulting approach remains faster than executing a full Newton-Raphson power flow solution for each operating point, while achieving much higher physical and operational fidelity than purely ML-based surrogates. Overall, PI-TTT enhances ML-based power flow solvers with improved physical and operational consistency while preserving their computational advantage, providing a promising foundation for further development towards real-time applications.

\begin{table}[t]
\centering
\setlength{\tabcolsep}{5pt}
\renewcommand{\arraystretch}{1.12}
\caption{Average per-sample runtime}
\label{tab:runtime}
\begin{tabular}{l l c}
\hline
\textbf{System} & \textbf{Model} & \textbf{Total Time (ms)} \\
\hline
IEEE 14   & PowerFlowNet            & 3.0 \\
          & PowerFlowNet + PI-TTT     & 17.0 \\
          & MF-GNN                    & 0.2 \\
          & MF-GNN + PI-TTT           & 17.2 \\
          & Newton-Raphson (NR)      & 19.0 \\
\hline
IEEE 118  & PowerFlowNet            & 3.0 \\
          & PowerFlowNet + PI-TTT     & 18.0 \\
          & MF-GNN                    & 0.2 \\
          & MF-GNN + PI-TTT           & 15.2 \\
          & Newton-Raphson (NR)      & 23.0 \\
\hline
IEEE 300  & PowerFlowNet            & 3.0 \\
          & PowerFlowNet + PI-TTT     & 18.0 \\
          & MF-GNN                    & 0.2 \\
          & MF-GNN + PI-TTT           & 15.2 \\
          & Newton-Raphson (NR)      & 29.9 \\
\hline
PEGASE 1354 & PowerFlowNet          & 3.0 \\
            & PowerFlowNet + PI-TTT   & 33.0 \\
            & MF-GNN                  & 0.2 \\
            & MF-GNN + PI-TTT         & 30.2 \\
            & Newton-Raphson (NR)    & 68.9 \\
\hline
\end{tabular}
\end{table}

\section{Conclusion}

In this study we introduce a physics-informed test-time training (PI-TTT) framework for enhancing the physical consistency and operational feasibility of ML-based power flow surrogates. By incorporating both AC power flow equations and operational constraints into a unified self-supervised objective, the method enables sample-wise refinement of surrogate predictions through a few gradient-based updates at inference time. The results across multiple benchmark systems demonstrate that PI-TTT substantially reduces power balance residuals and mitigates constraint violations compared with purely ML-based models, while retaining the computational efficiency that makes ML-based solvers attractive for real-time grid applications. The proposed approach effectively bridges the gap between fast but often inconsistent ML-based surrogates and the physically accurate solutions obtained from traditional numerical solvers. In small and medium-sized systems, PI-TTT produces operating points that are both physically feasible and operationally compliant. For larger networks such as the IEEE 300- and PEGASE 1354-bus grids, the method still yields notable improvements, although complete feasibility is not yet achieved. These results indicate that physics-informed refinement at inference time enhances the physical and operational consistency of ML-based power flow surrogates with a small additional computational overhead. Beyond improving the feasibility of these models, the approach can be also used as a lightweight refinement mechanism providing high quality warm starts for conventional Newton-Raphson solvers. With an additional computational budget or tailored optimization strategies, its benefits could extend further toward realistic large-scale operation.




\bibliographystyle{IEEEtran}
\bibliography{reference}
 
\end{document}